\documentclass[a4paper, 10pt, conference]{ieeeconf}      % Use this line for a4
\usepackage[left=54pt, right=37pt, top=122pt, bottom= 54pt]{geometry}

\usepackage{graphicx}
\usepackage{amsmath,amsfonts,amssymb}
\title{\LARGE \bf
Embedded Soft Sensing in Soft Ring Actuator for Aiding with the Self-Organisation of the In-Hand Rotational Manipulation}

\author{Ryman Hashem$^{1}$
and Fumiya Iida$^{1}$

\thanks{*This work was supported by the Engineering and Physical Sciences Research Council (EPSRC) RoboPatient grant EP/T00519X/1.}%

\thanks{$^{1}$ R. Hashem and F. Iida are with the Department of Engineering, University of Cambridge, UK
        {\tt\small rh805@cam.ac.uk}}%
        
}

\begin{document}

\maketitle
\thispagestyle{empty}
\pagestyle{empty}

%%%%%%%%%%%%%%%%%%%%%%%%%%%%%%%%%%%%%%%%%%%%%%%%%%%%%%%%%%%%%%%%%%%%%%%%%%%%%%%%
\begin{abstract}
This paper proposes a soft sensor embedded in a soft ring actuator with five fingers as a soft hand to identify the bifurcation of manipulated objects during the in-hand manipulation process. The manipulation is performed by breaking the symmetry method with an underactuated control system by bifurcating the object to clockwise or counter-clockwise rotations. Two soft sensors are embedded in parallel over a single soft finger, and the difference in the resistance measurements is compared when the finger is displaced or bent in a particular direction, which can identify the bifurcation direction and aid in the break of symmetry approach without the need of external tracking devices. The sensors performance is also characterised by extending and bending the finger without an object interaction. During an experiment that performs a break of symmetry, manipulated objects turn clockwise and counter-clockwise depending on the perturbation and actuation frequency, sensors can track the direction of rotation. The embedded sensors provide a self-sensing capability for implementing a closed-loop control in future work. The soft ring actuator performance presents a self-organisation behaviour with soft fingers rotating an object without a required control for rotating the object. Therefore, the soft fingers are an underactuated system with complex behaviour when interacting with objects that serve in-hand manipulation field. 

%We present a ring actuator with five fingers as a soft hand that manipulates objects with an underactuated system by breaking the symmetry method. The direction of the manipulated object bifurcates clockwise and counterclockwise. The bifurcation is achieved by changing the frequency, duty cycle and introduce perturbation to the system. We first present the concept of manipulating an object with five soft fingers in the ring actuator. A Simulink model that emulates the manipulation process is introduced, and empirical performance validation is conducted for simulation and actuator. An experimental platform is developed to investigate the rotational angles and the directions, which aided in analyzing the break of symmetry. We show that a soft underactuated system can manipulate objects by employing the soft material's morphology in the design. The fingers can extend and bend by applying pressure, and the object rotates as a result of the self-organization of the fingers without needing additional control signals. This method simplifies the control system in soft robotics manipulation by using the morphology of the robot's body.
%Go On with a summary of the next chapters

\end{abstract}

%%%%%%%%%%%%%%%%%%%%%%%%%%%%%%%%%%%%%%%%%%%%%%%%%%%%%%%%%%%%%%%%%%%%%%%%%%%%%%%%
\section{Introduction}

The soft robotics field has a significant impact on developing new methods for creating robots that were impossible to achieve with classical rigid methods \cite{rus2015}. With the increase of soft actuators, soft robots are usually tethered with many rigid parts as an actuation method, such as valves for pneumatic systems. While there is a trend to develop a soft actuation method like soft valves, minimising the tethered parts can be essential for some applications that require portability and low power consumption \cite{Ilievski2011, wehner2016}. Self-organisation is a concept that has been described the last few decades \cite{von1960}, though it has not been fully adapted in the soft robotics field \cite{pfeifer2012}. Underactuated soft robotics can be a good direction for dealing with multiple degrees of freedom to simplify the control system architecture. It can be possible to achieve an underactuated system by applying the first principles \cite{pfeifer2007}. Break of symmetry is a first principle method that can benefit soft robotics by analysing any bifurcation in the system that can be depicted and operated without majorly changing the control strategy of the system. The morphology of soft actuators is another aspect of self-organisation that can be investigated to achieve more adaptable behaviour without external control signals.

We propose a sensing method that can aid in the in-hand manipulation of rotating an object by embedding soft sensors on a finger in a soft ring actuator (RiSPA \cite{ryman2021} to have a self-sensing system with no external devices. RiSPA can manipulate the rotation of an object by breaking the symmetry method by applying a perturbation to the system with an underactuated technique. The in-hand manipulation in RiSPA is achieved with five soft fingers that rotate objects clockwise (CW) and counter-clockwise (CCW). The morphology of RiSPA is considered as an intelligence that can aid and reduce the control complexity. This method can be possible by using and distributing all parameters in the system as a coupled system and not dividing the morphology from the control scheme \cite{pfeifer2012}. The manipulation of the object is influenced by the self-organisation concept, where the fingers rotate in rhythmic sequence without a direct control interference. Thus, the ring actuator is an underactuated system that eliminates the control signal required for rotating an object. 

The challenges explored in this article are sensing a pneumatically actuated soft ring actuator with large deformation and classifying the manipulation process through the soft sensors. As the morphology and the behaviour of soft robotics are complex, this can lead to difficulties in designing and fabricating a soft sensory system \cite{dang2020sensorl}. In soft pneumatic robotics, the morphology of inflatable chambers is designed for a specific application such as manipulation \cite{Ilievski2011}. However, soft robots can serve different applications with the same morphology by changing the control architecture. The compliance of the materials in soft robotics is usually adaptive to the surrounding environment and can function accordingly \cite{mangan2005}. With these advantages of soft materials, an adaptive and compliant sensory system is essential for soft actuators.  

The stretchability of soft actuators introduces challenges for designing sensor techniques that adapt to soft materials behaviour and sense their stress and strain. With the large deformation produced by the ring actuator, a soft sensor that can adapt to this parameter without affecting the actuator performance is essential for the design. 

\begin{figure}[t]
    \centering
    \includegraphics[scale=0.95]{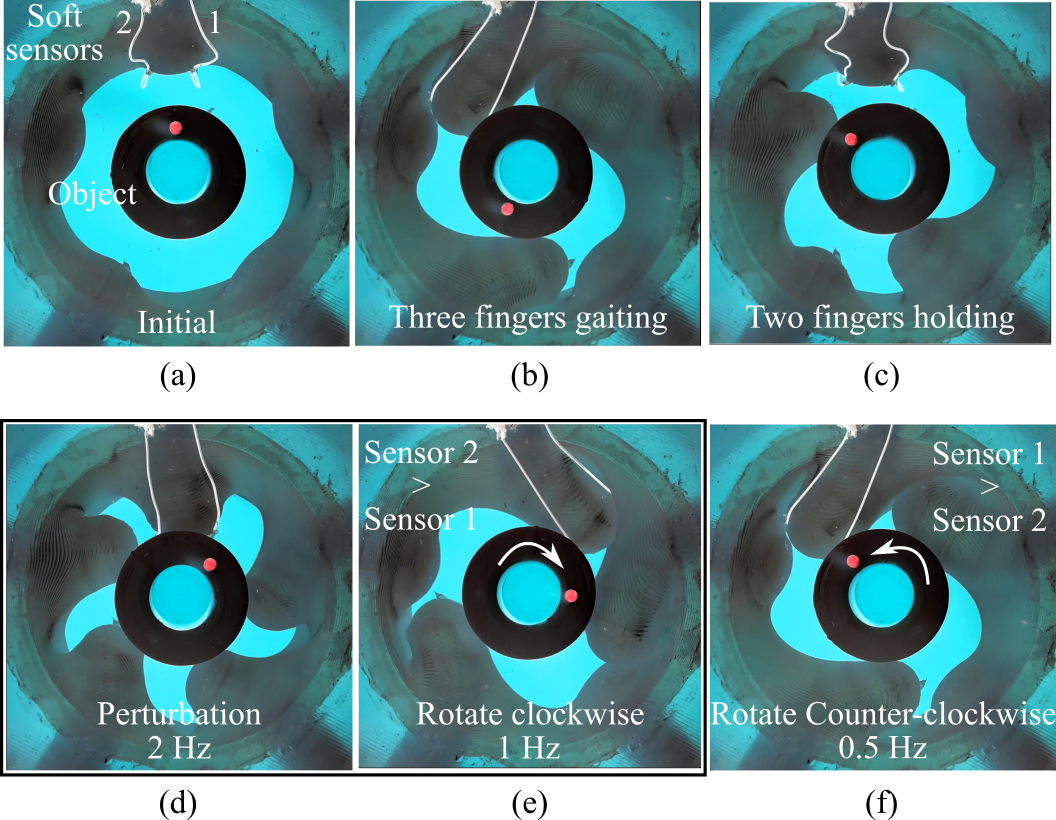}
    \caption{The concept of the gaiting of soft fingers to rotate an object with the self-organisation behaviour. (a) The initial position is when all fingers are fully retracted with -80 kPA of applied pressure. (b) The three rotating fingers object under a square signal of 100 kPa. (c) The two holding fingers for keeping the object from being influenced by the retracting rotating fingers. The concept of breaking the symmetry by applying perturbation and changing the frequency of applied pressure to change the direction of the manipulated object. (d) The perturbation of a single finger with an applied pressure of 90 kPa and a frequency of 2 Hz while the other four fingers are in a holding state under 50 kPa. (e) After the perturbation cycles end immediately, the gaiting process begins with a frequency of 1 Hz to rotate the object CW. (f) The rotation of object CCW without the perturbation and with a frequency of 0.5 Hz.}
    \label{fig2}
\end{figure}

Soft sensors are challenging to design and fabricate in soft robotics, and a custom design is usually proposed for a particular actuator. Large deformation is the challenge with the soft ring actuator, and few sensors can achieve that. Many soft actuators have been characterised by external tracking systems such as the Vicon and OptiTrack systems \cite{deng2016, ryman2021}. However,  External sensors are usually bulky, and they are not practical to use outside the validation process. We previously

installed an off-the-shelf proximity sensor (PS) and time-of-flight sensor (ToF) in RiSPA and swallowing robot, respectively, as a solution for measuring the displacement \cite{ryman2021,bhattacharya2021}. The advantages of the PS and ToF are in the small package size (they can be embedded inside a soft chamber) and the capability to detect a soft layer's deformation. These sensors cannot detect the bending motion because they require a clear space to perform the principle of the infrared sensing method. For the bending motion particularly, an embedded soft sensor is required to detect the bending motion. Embedded soft sensors are usually made of stretchable materials and conductive channels that measure the deformation by varying conductivity resistance. The stretchability of soft sensors is adaptable to soft actuators deformation without influencing the actuator's performance \cite{polygerinos2017}.

The conductive channels in soft sensors are usually filled with conductive fillers such as carbon black \cite{flandin2001}, carbon nanotubes \cite{lu2012}, liquid metals \cite{shin2014} or graphene \cite{kang2014}. Filled channels with additive materials measurement can discontinue as the channels can be pinched during deformation. A solution for this problem is by using a liquid as filling conductive such as eutectic gallium indium \cite{shin2014}. However, dealing with liquid additive increases the design and fabrication challenges and introduce contamination in case of leaking for some applications. Another method is a stain sensor to measure significant strain about 80\% by a mixture of thermoplastic elastomer (TPE) and carbon black particles \cite{mattmann2008}. The band is fibre-shaped with a diameter of 0.315 mm. Like the rubber band, this soft sensor can measure the displacement when stretched. This sensor can be applied easily on a soft actuator by defining the required deformation and how it can be measured. We adapted this soft sensor in this work as an embedded soft sensory system.

%\begin{figure}[t]
 %   \centering
  %  \includegraphics{Images/concept.png}
   % \caption{The ring-shaped soft pneumatic actuator (RiSPA) with the same morphological approach can serve various applications such as gripping soft objects, and simulating a segment of a human stomach or gastrointestinal tracts as presented in earlier work \cite{ryman2019,ryman2021}. RiSPA is employed here with another application of in-hand manipulation by an underactuated system with the self-organization and breaking-symmetry method. The fingers morphology comply with the cylindrical object and rotate it, and the arrow is for the illustration.}
   % \label{rispa}
  %\end{figure}

\begin{figure}[t]
    \centering
    \includegraphics[scale=0.96]{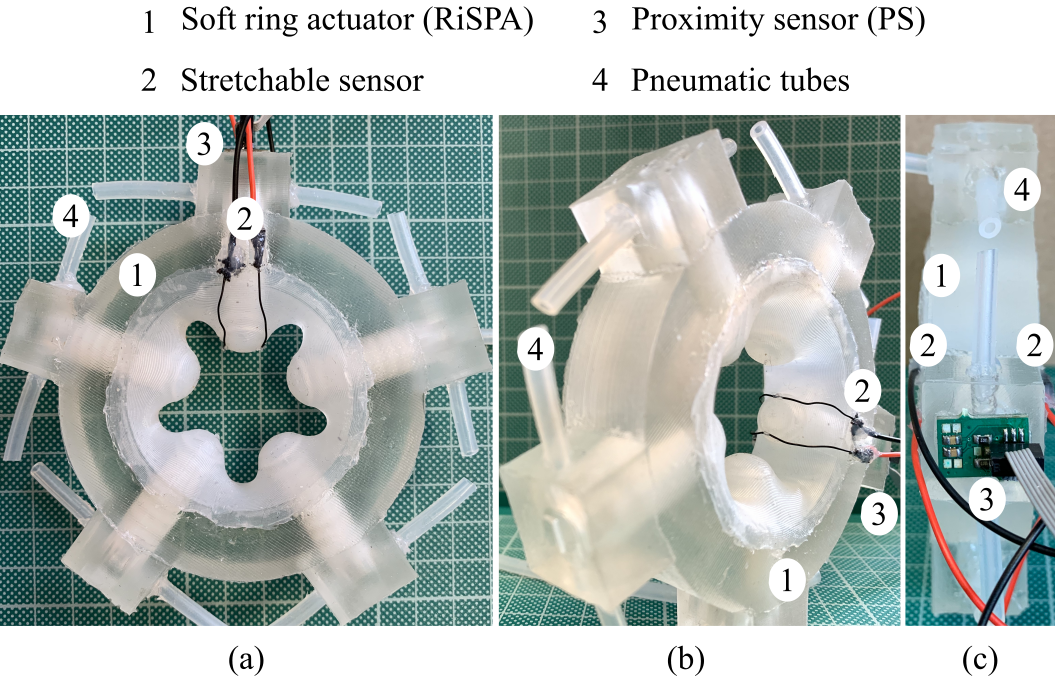}
    \caption{Soft ring actuator with three sensors installed, 2 soft sensors and one proximity sensor. The soft sensors are attached to the cap of the soft finger.}
    \label{fig1}
\end{figure}

In this article, a soft sensory system is adapted and embedded in RiSPA to aid in identifying the bending direction of the soft finger during the in-hand manipulation process. Firstly, we discussed the concept and setup of implementing soft sensors in RiSPA. Then, we presented experiments to validate the performance of the soft sensor in terms of displacement and bending without and with an object interacting with soft fingers. Finally, we performed a complete in-hand manipulation experiment while RiSPA rotating object CW and CCW to validate using a soft sensor with in-hand manipulation application without the need of external sensory systems.  The soft sensors showed promising results for detecting the direction of rotation and the dynamic deformation. The embedded soft sensors adapt to the soft actuator behaviour without disturbing the in-hand-manipulation functionality's performance.

%In this work, we focus on the rotational manipulation of cylindrical object shapes through the stabilization technique, restrain objects with two fingers while rotating with another three fingers \cite{bullock2012hand}. This paper presents a novel in-hand manipulation with a soft ring actuator that achieves complex behavior with a simple actuation method. We focus on the dexterity of motion at contact with objects. RiSPA has five embedded bellows actuators that are attached with a thin stretchable layer. The dimension of the actuator simulated a segment of the stomach. The number of actuators in the ring was also determined mathematically. The bellows actuators displace inward to the center of the ring to form a contraction (see Fig \ref{rispa}). An electro-pneumatic system provided the actuation. This paper explores another application that the soft ring actuator can achieve: the in-hand manipulation for robotics dexterity. We propose an underactuated soft actuator with the aid of break of symmetry behavior to reduce the challenges of precise control for dexterous manipulation. 

\section{Materials and method}

RiSPA was proposed earlier as a soft actuator that mechanically simulate a segment of the gastrointestinal tract \cite{ryman2021}. RiSPA has five soft fingers that displace toward the centre of the actuator frame to perform a contraction. When there is no interaction between the fingers and an object, the fingers extend linearly. However, in-hand manipulation occurs if an object is placed within the central section of RiSPA. If an object restrains the displacement of soft fingers, they accommodate the change by bending. While the contact between the soft finger and the object is intact, soft fingers influence and rotate the object while bending. To automate this behaviour, we proposed a gaiting method of having three soft fingers rotating the object. The other two fingers hold the object while the rotating fingers are retracted, so it does not rotate the object back to its position. The three fingers present a self-organisation behaviour, where they do not rotate against each other when the rotation direction is set to either CW or CCW. This method reduces the challenges of controlling the bending of soft fingers without an extra joint or complicated control system. Figure \ref{fig2} illustrates the gating method for manipulating an object with RiSPA.  

\begin{figure}[t]
    \centering
    \includegraphics[scale=0.9]{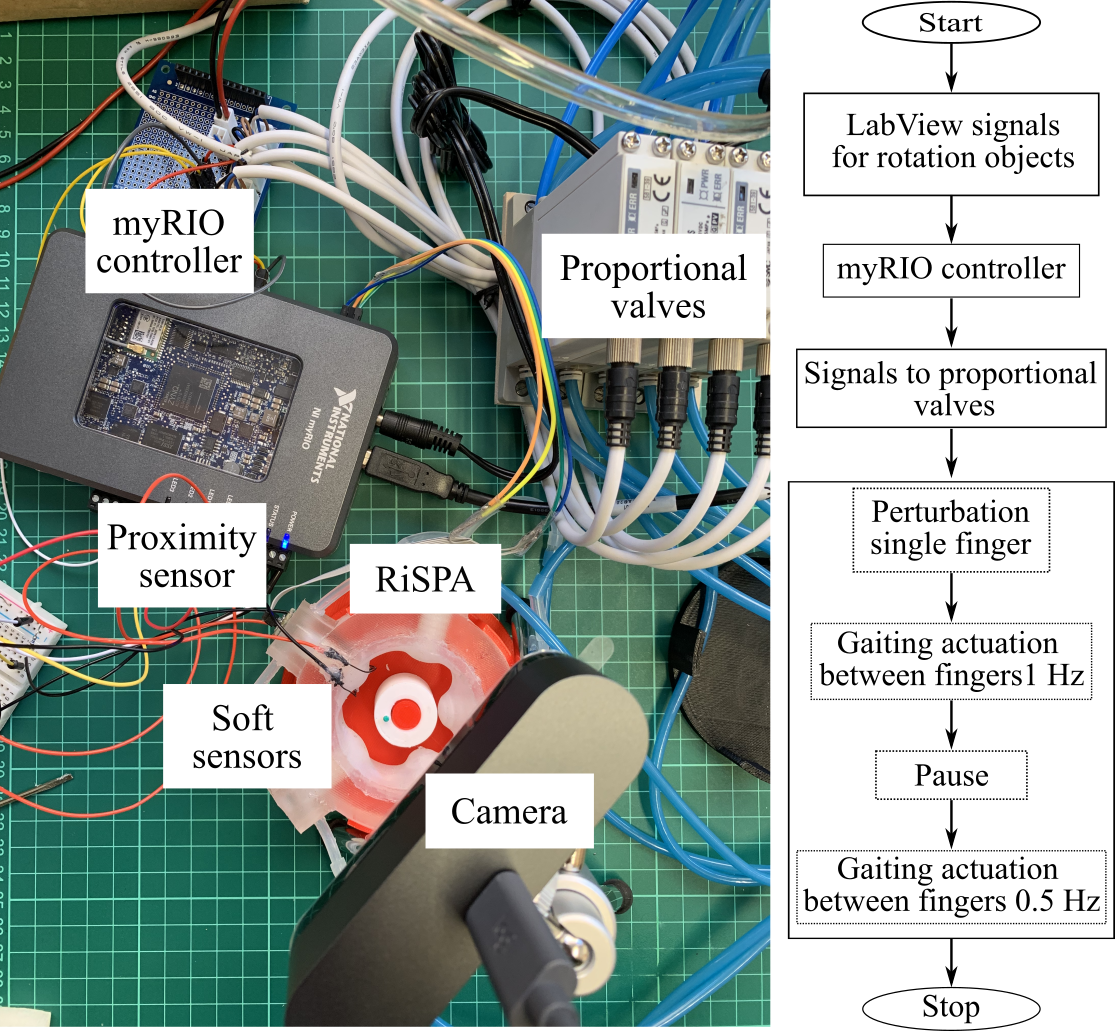}
    \caption{The experimental setup for characterising the soft sensors and tracking the rotation of the manipulated object and experimental architecture.}
    \label{fig3}
\end{figure}

The manipulation is performed in a cylindrical object and results in a rotational movement, either CW or CCW. The morphology of RiSPA aid in the control strategies as the system is underactuated. The morphology inherited self-organisation capability. The direction of rotation is performed by breaking the symmetry concept. The rotation of the manipulated object is dictated by the frequency of the applied signals and applying perturbation at the bifurcation point.

In this paper, we focus on embedding soft sensors to RiSPA to detect the bifurcation of the manipulation without external devices. The installation of soft sensor 1 (SS1) and soft sensor 2 SS2) is shown in figure \ref{fig1}. The soft sensors are placed in parallel, wrapping a cap of one of the soft fingers. We also used the PS (characterised in previous work \cite{ryman2021}) as a displacement sensor. The concept of detecting the bending motion of a soft finger is sown in figure \ref{fig2}. In the initial state, soft sensors have no change in measurements. When a manipulated object rotates CW by RiSPA, SS2 measurements is more significant than SS1; While rotating CCW results in SS1 $>$ SS2. From this, it is possible to analyse the performance of the in-hand manipulation system with the embedded self-sensing system.

\subsection{Experimental setup}
The setup for the conducted experiments is for characterising and validating the embedded soft sensors in RiSPA as shown in Figure \ref{fig3}. RiSPA and PS are accommodated from previous work \cite{ryman2021}. We used TPEs as soft sensors, which we can measure the resistance of the band when extended \cite{mattmann2008}. The soft sensors are glued at the cap of the soft fingers with silicone glue at a small point (sil-poxy, Smooth on, USA). The sensors are glued with the actuator is not pressurised, and the fingers are initially displaced up to 10 mm. Therefore, the soft sensors do not measure the retraction of the fingers from 0 to -80 kPa. The retraction process is not needed, and the PS sensor can track the displacement. 

For characterisation purposes of soft sensors, RiSPA was tested with no object interactions to measure the resistance of the soft sensors with displacement and bending. The displacement vs resistance test was performed by applying a step response from 0 to 100 kPa with 10 kPa intervals, and each step is 5 seconds long. Then, we selected a point from each step at 4.5 seconds to ensure the steady-state was reached for each step response. A dynamic test for the soft sensors was also performed by applying a sign wave with an amplitude of 100 kPa and frequency of 0.5 Hz. Bending behaviour was characterised by extending the soft finger with a fixed signal of 100 kPa, then forcing the soft finger with a lever of a stepper motor to bend from 0 to 25 degrees and 0 to -25 degrees. The stepper motor had an applied sine wave signal with a frequency of 0.5 Hz. 

\begin{figure}[t]
    \centering
    \includegraphics[scale=0.97]{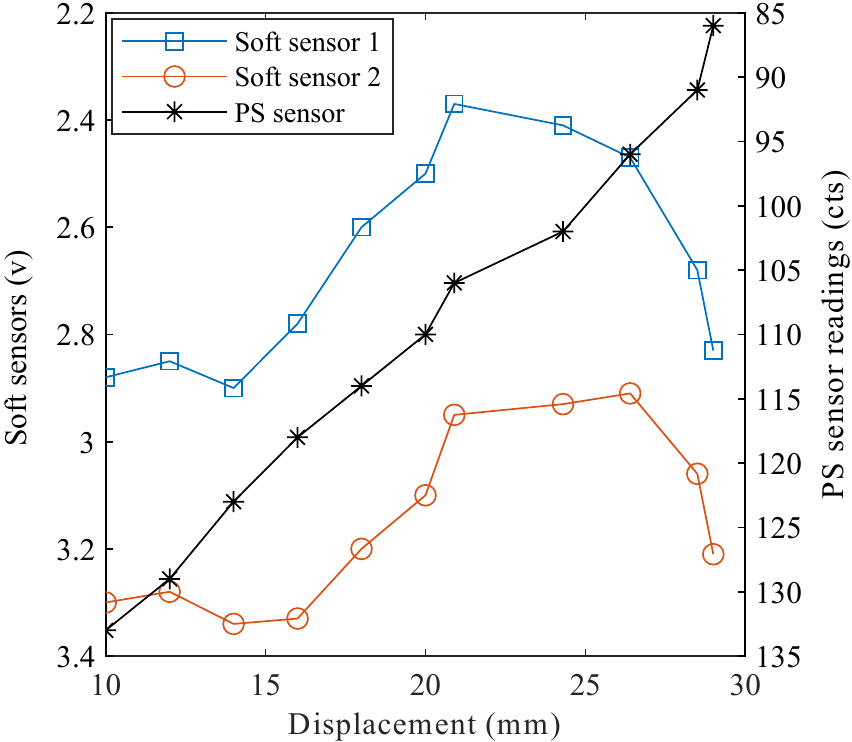}
    \caption{Sensor measurements against displacement of a single finger. }
    \label{fig7}
\end{figure}

The responses of the soft sensors during the in-hand manipulation process were analysed for the validation of detecting the manipulated object rotations. The applied control signals for this experiment is between -80 to 100 kPa of a square wave with a frequency of 1 Hz for rotating CW and 0.5 Hz for CCW. The setup for this experiment was performed with a 3D printed platform as shown in figure \ref{fig3} to only allows the rotation around the object axis. The object used is a cylindrical 3D printed part with a diameter of 22.5 mm. 

An electro-pneumatic system was used for applying pressurised air with five proportional valves (ITV0030, SMC Japan). We used  LabView software connected with myRIO for applying the control strategies (National Instrument).  We used an ACEIRMC SG90 Servo Motor for measuring the bending angle. We used a camera (Logitech brio) to track the object's rotation. MATLAB analysed the recorded videos to estimate the rotation of the objects. 

%We used electro-pneumatic systems to control the ring, each bellows connected to a proportional valve (ITV0030, SMC Japan). These valves are controlled by myRIO and LabView software (national instrument). Step function was used as applied signals with varied frequency, amplitude and phase delay between bellows actuators. The following equation illustrates the applied pressure. 

%The objects were tracked by a video camera (Lenovo) then the recordings were analysed by MatLab to abstract the rotational angles of the objects. The objects were designed with notches that were coloured for detection. 

%\begin{figure}[ht]
  %  \centering
  %  \includegraphics{Images/control-flowchart.pdf}
   % \caption{The control strategies for manipulating objects }
    %\label{FigureLabel1}
%\end{figure}

\begin{figure}[t]
    \centering
    \includegraphics{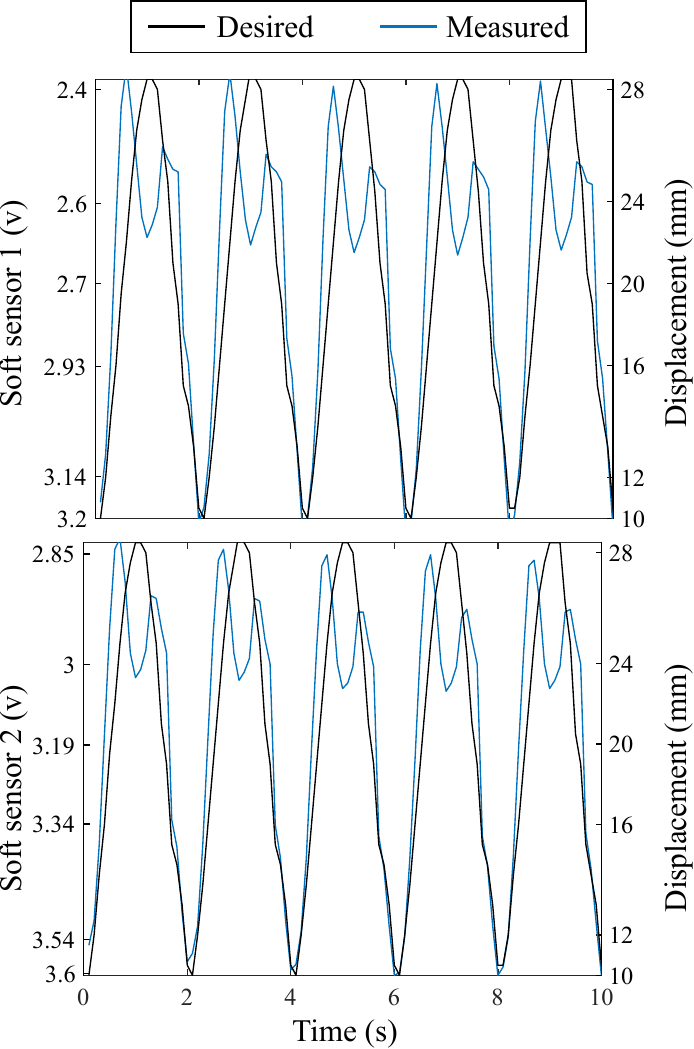}
    \caption{The dynamic responses of both soft sensors while the soft finger displaces by an applied pressure of sine wave with a frequency of 0.5 Hz and an amplitude of 100 kPa.}
    \label{fig8}
\end{figure}

\section{Results and discussion}
Figure \ref{fig7} shows the displacement responses of the soft sensors and the PS as abstracted from applied step pressure between 0 to 100 kPa. The displacement starts from 10 mm as the initial position of RiSPA when there is no vacuum or applied pressure. The retraction motion with a vacuum pressure is outside the scope of the paper. The responses of the soft sensors are nonlinear. When the sensors stretchability is reached to the maximum, sensors pull the soft layer attached to it. The increase of actuator displacement under higher pressure cause the soft sensors to retract. Soft sensors can measure a displacement of 9 mm on average, which emphasise that the PS sensor is still the best fit as a sensory system for the displacement measurement. However, the deflection, where soft sensors begin to retract, can be learned and adjusted by machine learning tools as it is a reasonably linear response. This behaviour showed the adaptation of the soft sensor in RiSPA even when the sensor stretchability was reached. Soft sensor behaviour does not affect the actuator's performance, and displacement of the soft finger is continued smoothly. 

Figure \ref{fig8} shows the dynamic behaviour of both soft sensors with an applied signal of a sine wave. The results presented a better performance than the static responses with step applied signals.  The sensors can follow the finger displacement with great response and repeatability. The scale of resistance of the dynamic response is also more significant than the static responses. During the retraction of the actuators, soft sensors have a convex shape behaviour. As explained in the static response of figure \ref{fig7}, the sensors pull the soft layer when they reach their maximum stretchability. The dynamic displacement trend does not get affected by the pulling behaviour. However, the descending deflation response presents the pulling effect with the concave shape. The overall response follows the displacement of the soft finger undergoing a cycle of a sine wave. Both embedded sensors have the same response behaviour.

\begin{figure}[t]
    \centering
    \includegraphics{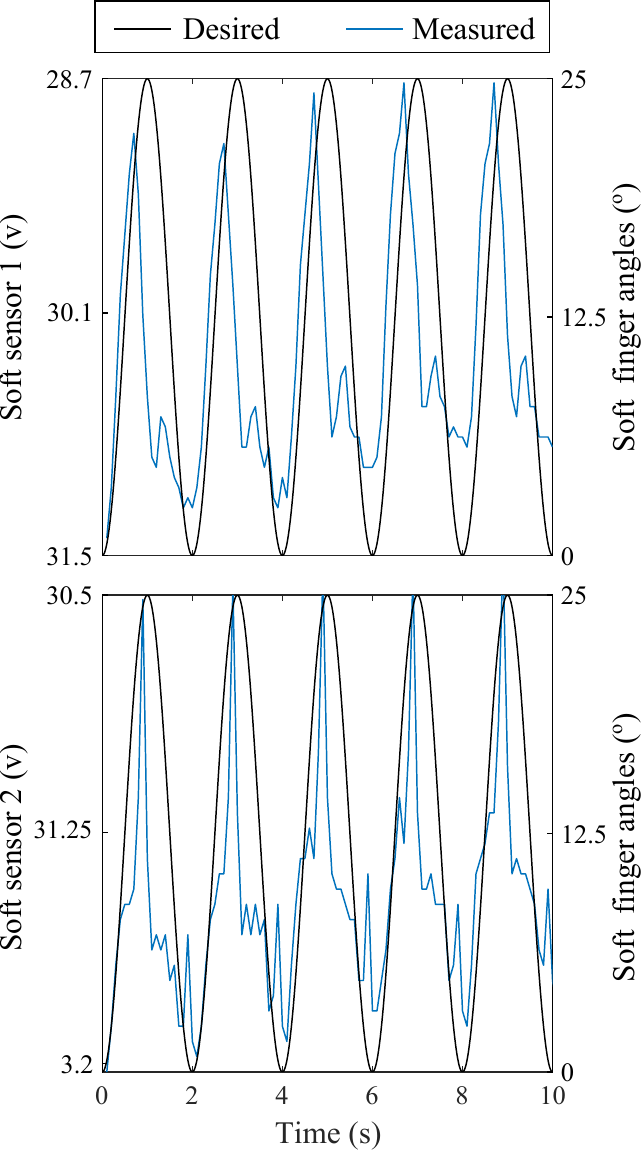}
    \caption{The response of bending the pressurised soft finger between 0 to 25 degrees. The soft sensors follow the desired bending angle with an average error of 5 degrees. When the fingers return to the initial position (0 bending angles), there is a noise in the reading of the senors.}
    \label{fig9}
\end{figure}

Figure \ref{fig9} illustrates the dynamic behaviour of a soft finger that has been forced with a servo motor with a sine wave to identify the response of the soft sensors. The soft sensors responses follow the desired bending angels. The retraction stage noises result from the contact between the servomotor lever and the soft finger. It is noted that bending the soft fingers to the left or right direction results in the same responses for both sensors. The PS sensor cannot detect any bending motion, and the response is static to any bending motion because PS functions as an infrared sensory system. The bending in the soft finger out of the PS detecting beam. Therefore, the embedded soft sensors are the ideal solution for bending motion. Combining the PS and the soft sensors presents a robust sensory system for RiSPA with detecting total displacement and bending motions. 

\begin{figure}[http]
    \centering
    \includegraphics[scale=1]{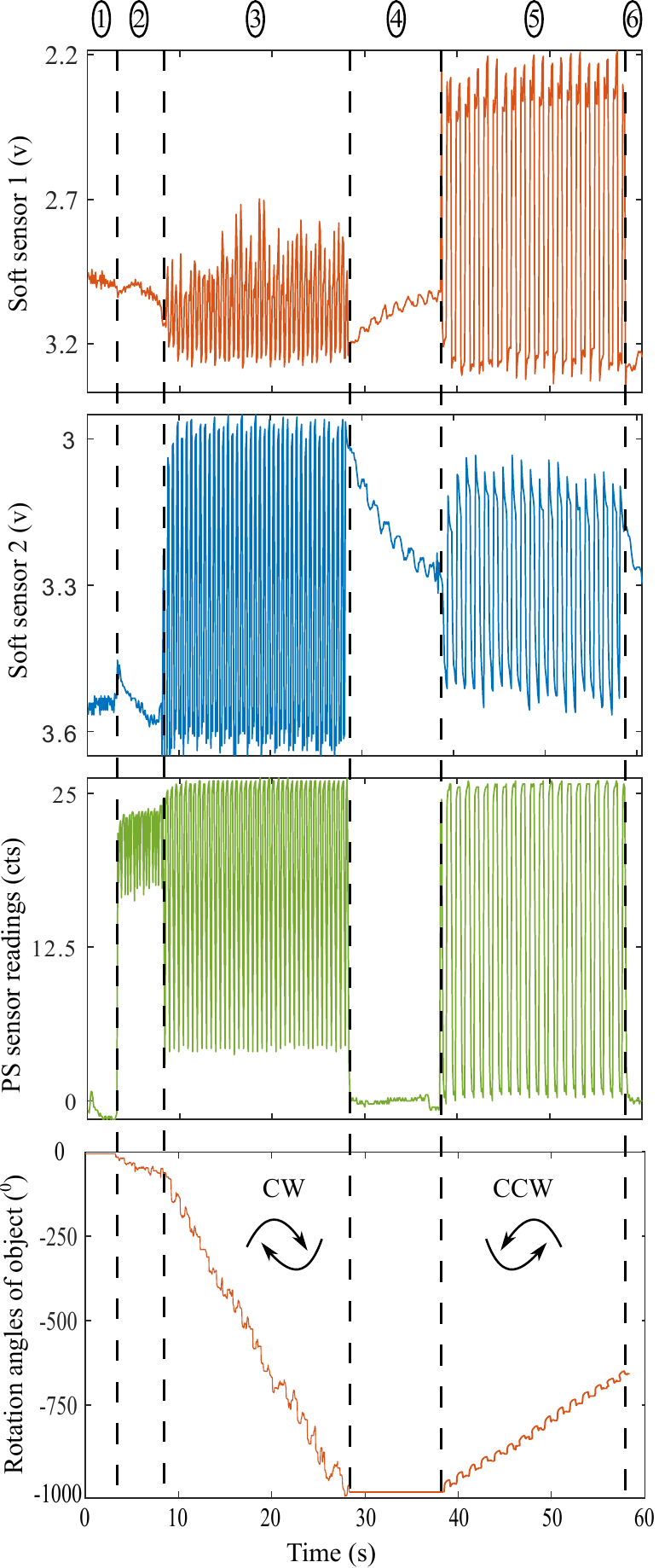}
    \caption{The responses of the three embedded sensors in RiSPA during the in-hand manipulation experiment presented the break of symmetry behaviour with a manipulated object as shown in the supplementary video S1. The strategy of the experiment is numbered from 1 to 6 as follows: initial position with fully retracted fingers with -80 kPa (1), the perturbation state with 90 kPa and a frequency of 2 Hz to break the symmetry (2), rotating the object CW with a frequency of 1 Hz and applied pressure of 100 kPa for the rotating fingers (3), the initial state with -80 kPa for all fingers (4), rotating the object CCW with a frequency 0.5 Hz and applied pressure of 100 kPa and without previous perturbation (5), initial state (6).}
    \label{fig10}
\end{figure}
%\subsection{Free axis platform}
%In this platform, the same control method of the fixed platform was applied to achieve both rotational directions. However, for the CW direction and where an initial actuation is required, a single actuator is actuated while the other four fingers are holding the object in the middle to restrict the movement of the objects. Unlike the the fixed platform, the object is not fully restricted with the four fingers and some displacements on the x-y plan occur, though the CW rotation is achieved. The behaviour of the initial single actuator between both platforms were diffident, but they resulted in a CW direction, which is the aim from this technique. 

Figure \ref{fig10} demonstrates the steps of the in-hand manipulation system with rotating an object CW and CCW. The three embedded sensors are compared during the test to identify the scale differences in the resistance measurements for the soft sensors.  The experiment started by fully retracting soft fingers. Then, the perturbation of a single finger with an applied pressure of 90 kPa and frequency of 2 Hz starts to break-the symmetry of the system and serve the manipulation with CW direction. During the perturbation and after 5 s, object manipulation with the gaiting technique begins with an applied pressure of 100 kPa for the rotating fingers and 50 kPa for the holding fingers with a frequency of 1 Hz. The holding fingers have a phase delay of 1$/$3 of the frequency. After 20 s, all soft fingers fully retract for 10 s. Then, object manipulation without perturbation and a lower frequency of 0.5 Hz begins, which rotates the object CCW. This manipulation also lasts 20 s to ensure repeatability and robustness of RiSPA. The video of the experiment is in the supplementary material S1. 

The PS responses show that the detection of an object in the system is achievable by the displacement measurement of the system. Comparing the applied pressure with the resulting displacement from the PS return different values when the actuator displaces without an object and when the object is in place. The free displacement with no object is about 28 under 100 kPa, while the displacement is under the same pressure when the object is in contact with the soft finger is 23 mm. This difference result from the bending motion where the soft finger appears shorter for the PS. However, the direction of bending cannot be detected with the PS. The detection of the object provides intelligence to the soft actuator and improve the in-hand manipulation quality by employing a closed-loop control when required.   

The behaviour of the soft finger while rotating an object to either CW or CCW is very stable and requires no external control. This behaviour demonstrates the self-organisation between the soft rotating fingers and the manipulated object. 

The rotational angle of the object shows the bifurcation behaviour clearly and the break of symmetry when the object undergoes the perturbation (rotate CW) or when there is no perturbation (rotate CCW). This phenomenon confirms the self-organisation inherited in the soft robot body and how it can be beneficial to simplify the control and design challenges. RiSPA as an in-hand manipulator can serve the field of soft robotics to manipulate an object with embedded sensors. Future work will focus on investigating a variety of object shapes and sizes to investigate 

\begin{figure}[ht]
    \centering
    \includegraphics[scale=0.93]{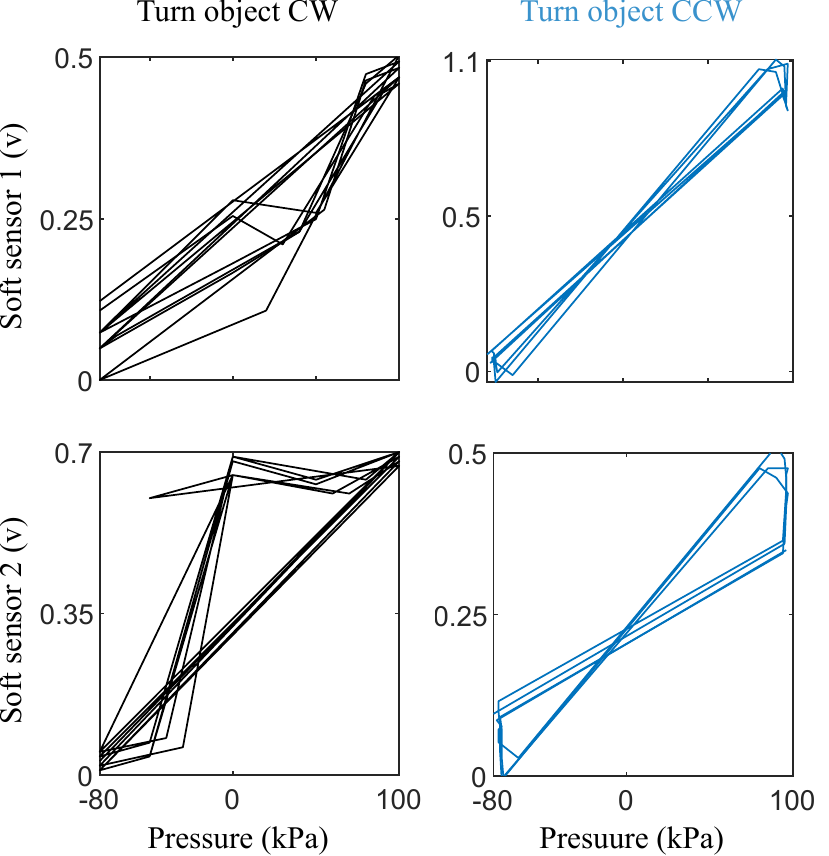}
    \caption{The measurements of resistance for soft sensor 1 and 2 during the in-hand manipulation process, pressure upon the resistance. The first column is the measurements of both sensors when object turning CW, while the second column is for the sensor measurements when object turning CCW. The differences in the measurement values are the comparing element between both sensors.}
    \label{fig11}
\end{figure}

Figure \ref{fig11} presents the results of the proposed method of two soft sensors embedded in a soft finger that can detect the rotation direction of the object when breaking the symmetry is achieved.  As proposed, both soft sensors' resistance varies depending on the direction of rotation with SS2 $>$ SS1 (0.7 $>$ 0.5) for turning an object with CW and SS1 $>$ SS2 (1.1 $>$ 0.5) for turning an object CCW. A single soft sensor is also can determine the direction with the difference of resistance. However, two sensors perform a robust and higher resolution for the detection. 

With embedded soft sensors in RiSPA, the in-hand manipulation can be fully automated without an external tracking device for mobile robotics applications.

\section{Conclusion}
We presented a method to sense the bifurcation of break of symmetry during the in-hand manipulating process with the soft ring actuator. We showed that the sensors detect the displacement and bending of the soft finger without interaction with an object. We also confirmed that the proposed method of embedding two soft sensors on the soft finger detects the manipulated object's bifurcation with either rotating clockwise or counter-clockwise. The soft ring system can measure the displacement, bending, detecting an object, and rotation direction. These sensing aspects are essential for autonomous in-hand manipulation applications. The soft fingers presented self-organisation behaviour with a repeatable response during the manipulation of the object where the rotation continued undisturbed without the need for external control signals. The intelligence of the actuator morphology simplified the control signal because of the adaptation of the soft fingers while interacting with objects.

%We demonstrated that the morphology of soft actuators aid in simplifying the control strategies of an underactuated system. A single input control broke the symmetry of the manipulation process by varying the applied pressure frequencies and duty cycles. A cylindrical object was used to demonstrate the concept of rotating clockwise and counter-clockwise with simple control. The flexibility and stretchability of soft fingers were modeled as prismatic and rotational joints to simulate the behavior of soft fingers. The simulation followed the actuator results closely in the counter-clockwise with 5\%, while a more significant deviation occurred on the clockwise scenario of 20\% because more tuning to the simulation was required, such as implementing dynamical friction to the object. This paper proposed a novel minimalistic experimental platform to introduce the self-organizing behaviors of in-hand manipulation systematically. The paper demonstrated the case when bifurcation happens in simulation and the real-world platform.

%The investigation of more manipulated objects with the aid of the simulation was left for future work. Also, different finger gating on manipulated objects is another aspect of analyzing the manipulation process. Another platform without object restriction is expected for the following study.

%%%%%%%%%%%%%%%%%%%%%%%%%%%%%%%%%%%%%%%%%%%%%%%%%%%%%%%%%%%%%%%%%%%%%%%%%%

%\section*{Acknowledgements} 

%\section*{Author Disclosure Statement}
%Detail any competing interests here.

%%%%%%%%%%%%%%%%%%%%%%%%%%%%%%%%%%%%%%%%%%%%%%%%%%%%%%%%%%%%%%%%%%%%%%%%%%

\bibliographystyle{ieeetr}
\bibliography{main.bib}

\end{document}